# The face-space duality hypothesis: a computational model

Jonathan Vitale (jonathan.vitale@student.uts.edu.au)
Mary-Anne Williams (mary-anne.williams@uts.edu.au)
Benjamin Johnston (benjamin.johnston@uts.edu.au)
University of Technology, Sydney
QCIS Centre - Innovation and Enterprise Research Lab, 15 Broadway - Ultimo NSW 2007

**Abstract**

Valentine's face-space suggests that faces are represented in a psychological multidimensional space according to their perceived properties. However, the proposed framework was initially designed as an account of invariant facial features only, and explanations for dynamic features representation were neglected. In this paper we propose, develop and evaluate a computational model for a twofold structure of the face-space, able to unify both identity and expression representations in a single implemented model. To capture both invariant and dynamic facial features we introduce the *face-space duality hypothesis* and subsequently validate it through a mathematical presentation using a general approach to dimensionality reduction. Two experiments with real facial images show that the proposed face-space: (1) supports both identity and expression recognition, and (2) has a twofold structure anticipated by our formal argument.

## Introduction

As an explanation of findings in face perception, Valentine used formal models of concept representations to propose that faces are represented in a psychologically plausible multidimensional space, *i.e.* the *face-space* (Valentine, 1991). Faces are points of this space based on their perceived properties. Valentine's formal models have been used to explain results of many human experimental studies, as well as computational simulations (Calder, Burton, Miller, Young, & Akamatsu, 2001; Lee, Byatt, & Rhodes, 2000; Rhodes et al., 2011).

These models were initially designed to only account for coding *identity-related* features, such as sex, distinctiveness, age and attractiveness (Calder et al., 2001). For example, the feature 'eyebrows' can vary from marked to delicate, thus possibly being one of the perceivable features crucial for coding the sex of a face. However, dynamic aspects of faces, such as facial expressions, were neglected. Early brain lesion and neuroimaging studies suggested that face identity and expression are not integral dimensions[1], instead they are represented and processed by separate systems that process faces in parallel (Bruyer et al., 1983; Tranel, Damasio, & Damasio, 1988). However, contemporary understanding indicates that identity and expression are more closely connected than previously thought, suggesting that common-codings can respond to both of them and so processes of their perception can interact (Ganel, Valyear, Goshen-Gottstein, & Goodale, 2005; Kadosh et al., 2016; Rhodes et al., 2015).

---
[1] Here and for the rest of the paper we use the term 'integral' to define properties exhibiting inter-dependencies and not completely separable.

Can Valentine's framework support an integral understanding of identity and expression processing? Calder and colleagues (2001) demonstrated that a multidimensional space derived from a principal component analysis (PCA) (Turk & Pentland, 1991) can provide a set of components being either identity-independent, expression-independent or identity-expression-interdependent (Calder & Young, 2005). However, in order to perform identity and expression recognition, the authors used two distinct latent discriminant analysis (LDA) modules, one choosing the best components to support identity recognition, whereas the other choosing the best components to support expression classification.

In this paper, we show that a *single* face-space with a twofold representation supports both identity and expression recognition. The structure of this face-space can be realized in a parsimonious way by integrating both identity and expression in a single model. We demonstrate the computational validity of our hypothesis through a rigorous mathematical presentation and related experiments. This work further supports the integral nature of identity and expression, at least from a computational perspective.

## Findings in Face Perception

Valentine's *'face-space'* framework (Valentine, 1991) is a notable cognitive model for face representation. According to this framework, facial representations are encoded in a multidimensional psychological space. The dimensions of this space are assumed to encode properties of the facial signals that better discriminate one face from another. The distance between two representations underlies their dissimilarity from a psychological perspective.

Identity and expression are two forms of facial information crucial for many social skills. Identity is considered an invariant feature of face, whereas expression a dynamic one. Traditional cognitive models of face perception suggest a complete separation of identity and expression systems after the completion of a structural encoding stage (Bruce & Young, 1986). Accordingly, Haxby *et al.* (2000) argue that invariant features, such as identity, and dynamic features, such as facial expression, are computed by separate regions of the brain.

However, new evidence from recent findings indicates that these systems operate interdependently (Pell & Richards, 2013). For example, Ganel and Goshen-Gottstein (2004) found that familiarity of faces increases the perceptual interdependence of identity and expression recognition. They suggested that differences between the facial configurations of individuals should lead to systematic differences in the way emotions are expressed by these individuals. For this reason, every individual can express each facial expression in a unique way. Knowledge of the identity of the observed subject can therefore facilitate the process of his or her facial expression.

We found a similar effect in developing a computational model inspired by simulation-theories (Vitale, Williams, Johnston, & Boccignone, 2014). In this endeavour, we suggested to first pre-process the observed face so to reduce its identity-related information, while preserving motor components information (*i.e.* its dynamic features). This facilitates the recognition of facial expression based on feed-forward mechanisms of internal motor simulation. A similar account was recently supported by a human study of Ipser and Cook (2015).

Calder *et al.* (2001) demonstrated the validity of integral identity and expression representations from a computational perspective. They submitted digital images of faces showing different identities and facial expressions to PCA thus obtaining representations based on components of a low-dimensional space. Their results demonstrated that this common representation can support both identity and expression recognition and that the representations of identity and expression partially overlap.

## The Face-Space Duality Hypothesis

In the previous section we provided studies supporting:

i A multidimensional spatial representation of faces as a plausible model for explaining many crucial effects in face perception;

ii An interdependence between representations of invariant and dynamic facial features;

iii That enhancing the separation between invariant and dynamic components of the face during face processing facilitates their classification.

Given this brief summary, we introduce the *'face-space duality hypothesis'*, suggesting that faces: (i) are encoded in a multidimensional face-space, (ii) under a common integral representation (iii) having a twofold structure: one supporting invariant features of the face (*e.g.* identity), whereas the other supporting dynamic features of the face (*e.g.* expression), thus contributing to their correct classification.

This hypothesis arises from the need to accommodate the apparently contrasting points (ii) and (iii) under a *single* representation based on a multidimensional space (i), which is an extension of the face-space framework proposed by Valentine (1991).

Consider a perceived face $\phi_i$ and a corresponding $d$-dimensional point $y_i$ of a multidimensional psychological space. The spatial representation $y_i$ encodes most of the original information of the input face $\phi_i$ and can be obtained through the mapping function $S(\phi_i) \mapsto y_i$. We introduce the functions $C^{\mathcal{E}}(\cdot) \to \mathbf{R}$ and $C^{\mathcal{I}}(\cdot) \to \mathbf{R}$, respectively providing the number of correctly classified facial expressions and identities from a set of observations encoding faces, and a permutation function $\sigma$ over the coordinates $(y^1, y^2, \ldots, y^d)$ of a point $y$ defined as:

$$\sigma = \begin{pmatrix} y^1 & y^2 & y^3 & \ldots & y^d \\ y^d & y^{d-1} & y^{d-2} & \ldots & y^1 \end{pmatrix} \quad (1)$$

We make use, here and for the rest of the paper, of the superscript $\tilde{\ }$ to denote a point or set of points to which is applied the permutation in (1). Then, given a set of perceived observations $\Phi$ and the associated multidimensional spatial representations $Y$, the mapping function $S$ is defined such that:

1. $S(\Phi) \mapsto Y$;

2. $C^{\mathcal{E}}(Y) \gg C^{\mathcal{E}}(\Phi)$;

3. $C^{\mathcal{I}}(\tilde{Y}) \gg C^{\mathcal{I}}(\Phi)$;

In other words, the face-space duality hypothesis assumes a function $S$ mapping the perceived faces onto a psychological multidimensional space having a twofold structure. This new representation allows the encodings $Y = \{y_1, y_2, \ldots, y_n\}$ and associated permutations $\tilde{Y} = \{\tilde{y}_1, \tilde{y}_2, \ldots, \tilde{y}_n\}$ to support significantly higher recognition rates than the original input representation $\Phi$.

The rationale behind this idea is that the dual face-space, by maximising the separation between dynamic and invariant features of the face in a single multidimensional representation, will order the components of the resulting space in such a way that the first ones will mostly correlate with dynamic features of the face, whereas the latter will mostly correlate with invariant features of the face. Therefore, the resulting face-space would provide a single multidimensional representation (as per point (i)) where invariant and dynamic features of the face are interdependent (as per point (ii)), but preserving a certain degree of separation able to support subsequent classification processes (as per point (iii)).

We investigate our hypothesis from a computational perspective, validating it using a mathematical analysis of a general dimensionality reduction framework used in face recognition, by limiting the analysis to facial identity and expression only. We further confirm our approach with experimental results.

## Dimensionality Reduction Models

The function $S$ introduced above can be modeled as a dimensionality reduction function. A dimensionality reduction function maps a high-dimensional signal onto a point of a low-dimensional space. For example, consider an image of a face having resolution $100 \times 100$ pixels. This observed signal is represented by a set of pixels and can be posed as a column vector $\phi_i$ of dimension $\mathcal{D} = 10000$. Dimensionality reduction models provide a mapping function $S : \mathbf{R}^{\mathcal{D} \times 1} \to \mathbf{R}^{d \times 1}$, with $d \ll \mathcal{D}$, such that the low-dimensional representation

$y_i = S(\phi_i)$ is able to explain the observed data $\phi_i$ (Yan et al., 2007).

Linear dimensionality reduction techniques make use of a linear projection matrix $V \in \mathbf{R}^{\mathscr{D} \times d}$ in order to map the high dimensional observed sample onto the low-dimensional target space. So the projection $y_i$ of an observation $\phi_i$ can be computed as $y_i = V^\top \phi_i$. When $V$ is an orthogonal matrix, an approximation of the original observation can be reconstructed from its projection: $\phi_i \approx V y_i$. The projection matrix $V$ can be estimated by solving an *objective function*. This objective function models desired constraints that the structure of the target low-dimensional space is required to satisfy.

For the purposes of this paper we limit our investigation to provide an implementation of our model through *linear* dimensionality reduction techniques.

## Graph-based dimensionality reduction framework

Yan *et al.* (2007) provided a general framework for unifying many dimensionality reductions models. Since our approach makes use of this framework, we first briefly introduce it below.

Let $\Phi = [\phi_i, \ldots, \phi_n]$ be a matrix of the $N$ observations represented as column vectors with dimension $\mathscr{D}$. The structure of the target low-dimensional space can be constrained by a similarity matrix $W$ and a penalty matrix $W^{(p)}$. For each pair of samples $(\phi_i, \phi_j)$, the similarity matrix $W_{ij}$ encodes the associated non-negative similarity measure, whilst the penalty matrix $W_{ij}^{(p)}$ encodes the associated penalty measure. This penalty measure can be used as a repulsive force between pairs of samples to prevent samples with high similarity but belonging to different classes from being placed in close proximity in the low-dimensional space (Kokiopoulou & Saad, 2009).

Given these two graph structures, the optimal mapping matrix $V^\star$ can be found by solving the following objective function:

$$V^\star = \underset{V \in \mathbf{R}^{\mathscr{D} \times d}}{\arg\min} \frac{Tr(V^\top \Phi L \Phi^\top V)}{Tr(V^\top \Phi L^{(p)} \Phi^\top V)} \quad (2)$$

with $Tr(\cdot)$ denoting the matrix trace operator and $L$, $L^{(p)}$ respectively being the resulting Laplacian matrices computed from the similarity and penalty matrices $W$ and $W^{(p)}$ (Yan et al., 2007).

Unfortunately, there is no closed-form solution to this optimization problem (Ngo, Bellalij, & Saad, 2012). However, the problem can be solved numerically with iterative algorithms whenever the matrix $\Phi L^{(p)} \Phi^\top$ is positive definite. The resulting optimal solution $V^\star$ is unique up to unitary transforms of the columns (Ngo et al., 2012).

To ensure that matrix $\Phi L^{(p)} \Phi^\top$ is positive definite, the process is usually split into two phases. First, given a dimension $\mathscr{D}' < N$, the observations $\Phi$ are provided in input to a PCA, and a first mapping matrix $V_{PCA} \in \mathbf{R}^{\mathscr{D} \times \mathscr{D}'}$ is estimated. Then, the samples $X = V_{PCA}^\top \Phi$ are provided as input to the objective function in (2) and the optimal mapping matrix $V^\star \in \mathbf{R}^{\mathscr{D}' \times d}$ is estimated. Finally, the overall mapping matrix able to reduce the dimensionality from dimension $\mathscr{D}$ to dimension $d$ and performing the constraints specified in the objective function (2) is given by:

$$V_{overall} = V_{PCA} V^\star \quad (3)$$

## Model implementation

Consider a set $\Phi$ of $N$ observations of frontal faces. Each observation consists of $\mathscr{D}$ pixel values, represented by a $\mathscr{D} \times 1$ vector. In this work we limit the investigation of observations varying only in identity and facial expression.

We set a dimension $d < N \ll \mathscr{D}$ and we estimate a mapping matrix $V_{PCA} \in \mathbf{R}^{\mathscr{D} \times d}$ by submitting the samples $\Phi$ to a PCA, thus obtaining the corresponding PCA-encodings $X = V_{PCA}^\top \Phi$. We aim to estimate another mapping matrix $V^\star \in \mathbf{R}^{d \times d}$, such that the final overall matrix $V_{overall} = V_{PCA} V^\star$ validates our hypothesis.

We denote the identity class of the sample $x_i$ with $\mathscr{I}(x_i)$ and the facial expression class of the sample $x_i$ with $\mathscr{E}(x_i)$.

Considering the previously introduced scenario, we start by designing the appropriate similarity and penalty matrices. Invariant features of the face extend to more regions of the face than dynamic ones, thus explaining most of the variance in the considered dataset of observations (Turk & Pentland, 1991). This means that during facial expression classification the identity can potentially introduce a bias on the samples (the *identity-bias*), thus increasing their similarity and reducing their distance in the face-space even when they belong to a different class of facial expression (Sariyanidi, Gunes, & Cavallaro, 2015).

Therefore, as a first step we design our similarity matrix to encourage pairs of samples associated with the same facial expression to be in close proximity in the resulting space, and our penalty matrix to provide a repulsive force between pairs of samples belonging to the same identity. This would result in maximising the separation between dynamic and invariant components of the face, thus facilitating the classification of facial expression.

Accordingly, we define the similarity matrix $W^{\mathscr{E}}$ as:

$$W_{ij}^{\mathscr{E}} = \begin{cases} \frac{1}{n_{\mathscr{E}_i}}, & \text{if } \mathscr{E}(x_i) = \mathscr{E}(x_j) \\ 0, & \text{otherwise.} \end{cases} \quad (4)$$

where $n_{\mathscr{E}_i}$ is the number of samples in $X$ belonging to facial expression class $\mathscr{E}(x_i)$.

Similarly, we define the penalty matrix $W^{\mathscr{I}}$ as:

$$W_{ij}^{\mathscr{I}} = \begin{cases} \frac{1}{n_{\mathscr{I}_i}}, & \text{if } \mathscr{I}(x_i) = \mathscr{I}(x_j) \\ 0, & \text{otherwise.} \end{cases} \quad (5)$$

where $n_{\mathscr{I}_i}$ is the number of samples in $X$ belonging to identity class $\mathscr{I}(x_i)$.

By using the proposed similarity and penalty matrices, the resulting Laplacians becomes $L = I_N - W^{\mathscr{E}}$ and $L^{(p)} = I_N - W^{\mathscr{I}}$, with $I_N$ a $N \times N$ identity matrix. Hence, these

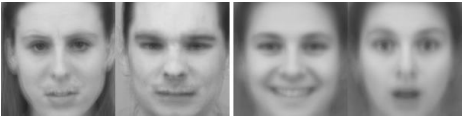

Figure 1: Some examples of prototypes. On the left are two prototypical identities (F05 and M07) in which expression-related features are reduced, whereas on the right are two examples of prototypical facial expressions (happiness and surprise).

Laplacians behave as block centring matrices. These matrices remove respectively the corresponding prototypical facial expression (*i.e.* an average identity showing the averaged facial expression) and the corresponding prototypical identity (*i.e.* the considered identity showing a neutral facial expression) from samples $X$ (see figure (1)).

The objective function in (2) becomes:

$$\arg\min_{V\in\mathbf{R}^{d\times d}} \frac{Tr(V^\top X(I_N - W^{\mathcal{E}})X^\top V)}{Tr(V^\top X(I_N - W^{\mathcal{I}})X^\top V)} \quad (6)$$

We solve the objective function in (6) with the iterative algorithm proposed in (Ngo et al., 2012). Given the matrices $M^{\mathcal{E}} = X(I_N - W^{\mathcal{E}})X^\top$ and $M^{\mathcal{I}} = X(I_N - W^{\mathcal{I}})X^\top$ the optimal mapping matrix $V^\star$ can be found through the algorithm (1).

---

**Data**: Matrices $M^{\mathcal{E}}$, $M^{\mathcal{I}}$, a maximum number of iterations $K$ and a tolerance $\varepsilon$.
**Result**: A mapping matrix $V$ of dimension $D \times d$.
$V \leftarrow I_{D\times d}$;
**for** $i \leftarrow 1$ **to** $K$ **do**
 $\rho \leftarrow \frac{Tr(V^\top M^{\mathcal{E}} V)}{Tr(V^\top M^{\mathcal{I}} V)}$;
 $\mathcal{G}(\rho) \leftarrow M^{\mathcal{E}} - \rho M^{\mathcal{I}}$;
 Compute the smallest (for minimisation) or largest (for maximisation) $d$ eigenvalues $[\lambda_1,\ldots,\lambda_d] \equiv \Lambda$ of $\mathcal{G}(\rho)$ and associated eigenvectors $[v_1,\ldots,v_d] \equiv V$;
 **if** $|\sum_{j=1}^{d}\Lambda| < \varepsilon$ **then**
  | break;
 **end**
**end**

**Algorithm 1:** Newton-Lanczos algorithm for optimization of objective function (6).

---

From the algorithm (1), it can be seen that the optimal mapping matrix $V^\star$ is the set of eigenvectors associated with the smallest eigenvalues of $\mathcal{G}(\rho^\star)$, with $\rho^\star$ being the result of the trace ratio in (6) when posing the optimal solution $V^\star$. If, instead of taking the eigenvectors associated with the smallest eigenvalues, we take the eigenvectors associated with the largest eigenvalues, we get the optimal solution for the following objective function:

$$\arg\max_{V\in\mathbf{R}^{d\times d}} \frac{Tr(V^\top X(I_N - W^{\mathcal{E}})X^\top V)}{Tr(V^\top X(I_N - W^{\mathcal{I}})X^\top V)} \quad (7)$$

Using simple properties of trace and eigenvalues, it follows that the objective function in (7) can be equivalently posed as:

$$\arg\min_{V\in\mathbf{R}^{d\times d}} \frac{Tr(V^\top X(I_N - W^{\mathcal{I}})X^\top V)}{Tr(V^\top X(I_N - W^{\mathcal{E}})X^\top V)} \quad (8)$$

By reminding that a centring matrix is symmetric idempotent and $Tr(AA^\top) = \|A\|_2^2$, it is possible to note that the objective function (6) attempts in minimising the distances between the encodings $Y = V^\top X$ and their corresponding prototypical facial expressions $Y_{proto}^{\mathcal{E}} = V^\top X W^{\mathcal{E}}$, while maximising their distances with respect to the prototypical identity $Y_{proto}^{\mathcal{I}} = V^\top X W^{\mathcal{I}}$, overall facilitating expression recognition. Conversely, the objective functions (7, 8) attempts in minimising the distances between the encodings $Y = V^\top X$ and their corresponding prototypical identities $Y_{proto}^{\mathcal{I}} = V^\top X W^{\mathcal{I}}$, while maximising their distances with respect to the prototypical facial expression $Y_{proto}^{\mathcal{E}} = V^\top X W^{\mathcal{E}}$, overall facilitating identity recognition.

Since the eigenvectors of $V^\star$ are estimated from the same matrix $\mathcal{G}(\rho^\star) = M^{\mathcal{E}} - \rho^\star M^{\mathcal{I}}$ in both the objective functions (6,7), the components of the two spaces are the same, but differing by order. In other words, given $V^\star$ as the optimal matrix resulting from objective function (6) we can easily get the optimal matrix of the objective function (7) $\tilde{V}^\star$, defined as a matrix with the same columns of $V^\star$, but arranged in an inverse order.

Finally, given the matrix $V_{PCA} \in \mathbf{R}^{\mathcal{D}\times d}$ and the matrix $V^\star \in \mathbf{R}^{d\times d}$ we can estimate the final mapping matrix $V_{overall}$ of the face-space through equation (3), which leads respectively to the mapping $Y = V_{overall}^\top X$ and the associated permutated mappings $\tilde{Y} = \tilde{V}_{overall}^\top X$ as suggested by our hypothesis.

Note that we are not claiming that this is the only way to implement the proposed mapping function $S$ (for example it can be generalised to non linear models), and neither that human brain implement the suggested dual face-space in this way. However, this is a viable computational implementation of the proposed model able to support our hypothesis. In the remainder of this paper we further support the face-space duality hypothesis with experimental data.

## Experiments

We further validate our hypothesis using images from the Karolinska Directed Emotional Faces (KDEF) dataset (Lundqvist, Flykt, & Öhman, 1998). The dataset contains static images of 70 subjects—35 female and 35 male—exhibiting 7 different prototypical facial expressions of basic emotions (anger, disgust, fear, happiness, neutral, sadness and surprise). The pictures are taken in different face orientations and in two different sessions (A and B).

We used the frontal pictures taken in session A. The facial region was extracted from the images and its resolution reduced to 80 × 80 pixels. Eyes and mouth were at approximately the same position. Illumination variations were reduced by applying a simple equalization process to the images.

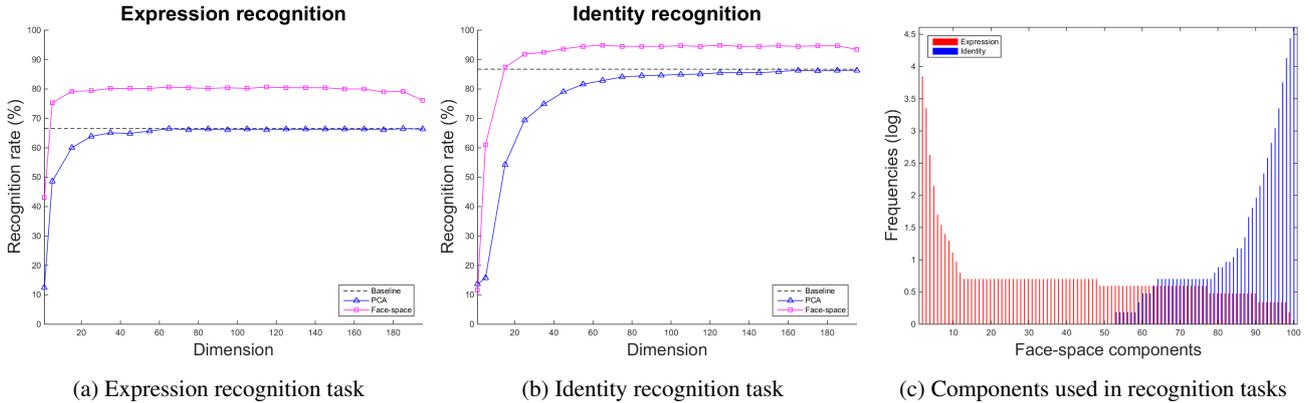

(a) Expression recognition task  (b) Identity recognition task  (c) Components used in recognition tasks

Figure 2: Results of the proposed experiments.

We first pre-processed the data by submitting the pixels of the images in input to a PCA as explained previously. In the first experiment we retained the components able to explain 95% of the variance of the original data resulting in 200 components, while in the second experiment we retained the data explaining the 85% (so reducing the number of components and allowing better readability) resulting in 100 components.

We performed two experiments: the first to test the ability of the proposed face-space in supporting identity and expression recognition, and the second to demonstrate the twofold nature of the resulting face-space.

**Support of identity and facial expression recognition**

In the first experiment we test the ability of our model to support subsequent processes of identity and facial expression recognition. We used a 10-fold cross validation approach. For each iteration we divided the data by taking one fold as test and the rest for model training.

With each training data we estimated the mapping matrix $V_{overall}$ as per equations (6, 3). Then we mapped each test data onto the corresponding face-space, thus obtaining the encodings $Y^{\mathcal{E}} = V_{overall}^\top \Phi$ and $\tilde{Y}^{\mathcal{I}} = \tilde{V}_{overall}^\top \Phi$ respectively used during the expression and identity recognition tasks.

The classification was performed using the nearest neighbour algorithm. For each sample $\phi_i$, $x_i$ and $y_i$ we computed the Euclidean distances with respect to the centroids of each class in the corresponding space (i.e. the prototypical identities in the case of identity recognition or the prototypical expressions in the case of facial expression recognition) and selected the label associated with the centroid having lower distance to the sample. We repeated this process for each dimension $k$ by taking only the first $k$ components of the encodings $x_i$ and $y_i$. In classifying the raw observations $\phi_i$ for a baseline comparison, we considered all the pixel values of the input images as coordinates of points in a $\mathcal{D}$-dimensional space. The recognition rates in each dimension were averaged among each cross-validation test.

The results for facial expression and identity recognition are shown, respectively, in figures (2a) and (2b). It is clear that this face-space derived by a single integrated process can support both identity and expression recognition, whereas a simple PCA cannot overcome the baseline performance.

**Validation of face-space twofold structure**

We were able to confirm our hypothesis on the twofold structure of the face-space using data.

We estimated the mapping matrix $V_{overall}$ as per equations (6, 3) using the full dataset as training data.

Given the matrix $V_{overall} = [v_1, \ldots, v_d]$ the minimum set of expression components for a sample $\phi_i$ is the smallest set $V_{i_{min}} = [v_1, \ldots, v_k]$ such that $y_i = V_{i_{min}}^\top \phi_i$ is classified with the correct expression label $\mathcal{E}(\phi_i)$ through nearest neighbour algorithm, as in the previous experiment.

Similarly, given the matrix $\tilde{V}_{overall} = [v_d, \ldots, v_1]$ the minimum set of identity components for a sample $\phi_i$ is the smallest set $\tilde{V}_{i_{min}} = [v_k, \ldots, v_1]$ such that $y_i = \tilde{V}_{i_{min}}^\top \phi_i$ is classified with the correct identity label $\mathcal{I}(\phi_i)$ through nearest neighbour algorithm, as in the previous experiment.

For each sample $\phi_i$ we computed its minimum set of expression and identity components. Then, we set $n_k^{\mathcal{E}}$ the number of times the component $k$ was included in the minimum sets of expression components and $n_k^{\mathcal{I}}$ the number of times the component $k$ was included in the minimum sets of identity components. We computed the final results for expression and identity and for each component $k$ as $f_k^{\mathcal{E}} = \log(\frac{n_k^{\mathcal{E}}}{N} \times 100 + 1)$ and $f_k^{\mathcal{I}} = \log(\frac{n_k^{\mathcal{I}}}{N} \times 100 + 1)$. Here $N$ is the number of samples in the dataset (i.e. 490) and we used the logarithm for better readability of the results. The resulting log-frequencies are illustrated in Figure (2c).

From the results two peaks placed in the extremes of the face-space components are clearly visible. There are expression components clearly independent from identity ones (components #1 to #52), components shared among expression and identity classification tasks (components #53 to #99) and just one identity component independent from expression ones (component #100). These results are in agreement with the study of Ganel and Goshen-Gottstein (2004), which suggest that expression is perceptually separable from identity, but identity is not perceptually separable from expression. This experiment further supports the proposed twofold structure of face-space as suggested by the face-space duality hypothesis.

## Conclusions

In this paper we extended Valentine's face-space framework to better explain recent findings in face perception.

In alignment with recent studies in face processing, we suggested that the nature of identity and expression dimensions is highly interdependent. We proposed that the structure of the face-space can result from a single process integrating invariant and dynamic features of the face. From this process results a twofold structure. Reading the components of this space from the first to the last facilitates classification of dynamic features of the face, such as facial expression, while reading them from the last to the first better supports classification of invariant features, such as the identity.

We validated the face-space duality hypothesis, from a computational perspective, through a formal mathematical presentation, by considering a general framework of dimensionality reduction. The hypothesis was further supported by experimental data.

This framework might serve as the basis and inspiration for future empirical work on face processing and linked to neural approaches.